\documentclass{article}

\usepackage{arxiv}

\usepackage[utf8]{inputenc} % allow utf-8 input
\usepackage[T1]{fontenc}    % use 8-bit T1 fonts
\usepackage{hyperref}       % hyperlinks
\usepackage{url}            % simple URL typesetting
\usepackage{booktabs}       % professional-quality tables
\usepackage{amsfonts}       % blackboard math symbols
\usepackage{nicefrac}       % compact symbols for 1/2, etc.
\usepackage{microtype}      % microtypography
\usepackage{lipsum}		% Can be removed after putting your text content
\usepackage{graphicx}
\usepackage{natbib}
\usepackage{doi}
\usepackage{amsmath}
\usepackage{booktabs}
\usepackage[most]{tcolorbox}
\usepackage{listings}
\tcbuselibrary{breakable, skins, listings}
\usepackage{color}
\usepackage{caption}
\usepackage{graphicx}
\usepackage{subcaption}

\title{A Hybrid Agentic AI Framework for Intelligent Supply Chain Analytics}

\author{
Xian Yeow Lee, Teppei Inoue, Haiyan Wang, Chetan Gupta \\
Industrial AI Lab \\
Hitachi America \\
\texttt{xian.lee@hal.hitachi.com, teppei.inoue.wa@hitachi.com} \\
\texttt{haiyan.wang@hal.hitachi.com, chetan.gupta@gmail.com}
}

\date{}

\renewcommand{\shorttitle}{A Hybrid Agentic AI Framework for Intelligent Supply Chain Analytics}

\hypersetup{
pdftitle={A Hybrid Agentic AI Framework for Intelligent Supply Chain Analytics},
pdfsubject={Supply Chain Analytics, Agentic AI, Industrial AI},
pdfauthor={Xian Yeow Lee, Teppei Inoue, Haiyan Wang, Chetan Gupta},
pdfkeywords={Agentic AI, Supply Chain Analytics, Intelligent Supply Chain, Hybrid AI, Industrial AI},
}

\begin{document}
\maketitle

\begin{abstract}

Efficient utilization of supply chain analytics for decision making remains a significant challenge for planners, as critical tasks such as database querying, key performance indicator (KPI) analysis, demand forecasting, and performance diagnosis require heterogeneous expertise spanning data engineering, operations research, and domain knowledge. In this work, we propose an agentic system for supply chain analytics that bridges the gap between business decision-making and technical expertise, where a coordinator agent interprets user intent and delegates sub-tasks to specialized agents. The system supports both exploratory analysis and deterministic workflows, enabling planners to transition between ad hoc questions and structured processes. Domain logic is encapsulated within specialist agents and prompts, yielding a scalable, modular, and auditable design and lowering the cost of functional extension through prompt-centric development. We evaluate the proposed architecture on a test environment that replicates multi-echelon inventory management operations. Results show that our multi-agent design achieves a 90\% accuracy, which is competitive with a single agent baseline while reducing input token usage by roughly fourfold, substantially improving scalability and cost-efficiency. Furthermore, we provide case studies to demonstrate interpretable suboptimality detection and automated forecast optimization, illustrating how agentic architectures can effectively combine  open-ended exploratory analysis  and deterministic supply chain analytics workflows, and provide a practical pathway toward more accessible and extensible decision-support systems.

\end{abstract}

%TODO
%Compare single and multi-agent, as a data point to the community
%Support multi-agent, but limitations
%Include tooling at every agent

%Benefits:
%Need to include the fact of token usage of multi-agent is also lower since we do not carry the entire prompt
% A couple sentences on reasoning vs non-reasoning in results since we remove from related works
%Citations
%Add samples
%%%%%%%%%%%%%%%%%%%%%%%%%%%%%%%%%%%%%%%%%%%%%%%%%%%%%%%%%%%%%%%%%%%%%%%%%%%%%%%%

\section{INTRODUCTION}

Modern supply chain management (SCM) systems require the integration of diverse analytical capabilities, including database querying, key performance indicator (KPI) computation, performance analysis, demand forecasting, and optimization. In practice, these capabilities are supported by heterogeneous tools and are often owned by different organizational roles. Transactional and operational data are typically stored in relational databases with company-specific schema, while performance assessment relies on a multi-dimensional set of domain-specific KPIs, with different industries emphasizing different performance metrics, such as days inventory outstanding (DIO) and fill rate in retail supply chains. More advanced tasks, such as sub-optimality detection or forecast refinement, further require domain knowledge, modeling assumptions, and structured analytical workflows. As a result, expertise is fragmented across data engineers, supply chain analysts, and operations researchers, creating friction for planners who require timely and actionable data-driven insights.

At the same time, planners increasingly need to combine ad hoc exploratory analysis, such as diagnosing why a particular item under-performs at a subset of stores, with standardized, repeatable workflows such as statistical and machine learning based demand forecasting. Existing SCM systems typically force a trade-off between these needs. Deterministic workflow engines provide reliable and auditable outputs but require users to operate through rigid interfaces and predefined parameters. In contrast, recent generative AI-based assistants enable natural-language interaction and open-ended exploration, but often lack domain safeguards and schema knowledge and struggle to incorporate structured workflows. Consequently, a unified system with a natural-language interface that supports both flexible exploration and domain-safe execution remains largely absent in operational SCM settings.

From a SCM system perspective, extensibility and maintainability further complicate this landscape. Supply chain analytics evolve continuously as new KPIs, workflows, and data sources are introduced. Traditional systems frequently hard-code logic into application code, making even minor changes costly and slow to deploy. Large language model (LLM)-based systems with tool-calling capabilities encode analytical logic, routing decisions, and workflow structures in agent instructions rather than in application code, as in traditional systems, thereby enabling rapid iteration and lowering adoption barriers.

However, LLM-based systems face inherent scalability and clarity limitations, especially for industrial applications. A single monolithic agent that is expected to perform all the diverse tasks must operate under long and complex instructions, leading to context pressure, brittle behavior, and limited interpretability. Conversely, purely pre-defined workflow engines lack the flexibility required for open-ended analysis and early-stage diagnosis. Existing approaches therefore struggle to balance flexibility with control, and exploration with safety. In this paper, we propose a multi-agent, hub-and-spoke architecture for natural-language supply chain analytics that addresses these limitations. The core contributions of this work are:

\begin{itemize}
    \item \textbf{Hybrid agentic architecture:} We introduce a framework that combines domain-specific deterministic workflows with open-ended agentic analysis, enabling both safe execution of structured supply chain processes and flexible exploratory reasoning within the same system.
    \item \textbf{Scalable multi-agent design:} We present a coordinator-centric, hub-and-spoke architecture in which specialized agents handle distinct analytical responsibilities. This design improves scalability, auditability, and effective context usage compared to single-agent baselines.
    \item \textbf{Evaluation and use cases:} We evaluate and compare the performance of a single agent system with a multi-agent system and demonstrate the proposed architecture through a concrete supply chain use case, evaluated using a proprietary industry simulator that acts as a supply chain digital twin.
\end{itemize}

Together, these contributions illustrate how hybrid agentic systems can provide a practical and extensible foundation for supply chain analytics, bridging the gap between exploratory natural-language interaction and deterministic, domain-specific decision workflows.

\section{Related Work}

\subsection{LLMs with Tool Use and Agentic Systems}

Recent advances in LLMs enable natural-language interaction with external tools such as databases and code execution environments. Early text-to-SQL work established methods for translating queries into structured database commands~\cite{nlp_2_sql1, nlp_2_sql2}, with later LLM-based approaches improves schema and conversational generalization~\cite{llm_2_sql1}. These surveys summarize progress in semantic parsing and natural-language database interfaces~\cite{nlp_2_sql_survey, llm_2_sql_survey}. Beyond query translation, tool-augmented LLMs support function calling and multi-step workflows~\cite{agentic_workflow_survey1}, but monolithic agents that combine querying, reasoning, and tool-calling often face scalability and context limitations. To address this, multi-agent architectures delegate subtasks to specialized agents under centralized coordination~\cite{Sarkar2025SurveyOLA}, with applications in automation, scientific discovery, and forecasting~\cite{Hu2025OWLOWA, Pantiukhin2025AcceleratingESA, Zuo2025LargeLMA, herron2025hierarchical}. 

\subsection{Intelligent Supply Chain Management}

Supply chain management has long leveraged quantitative models for forecasting, inventory control, and performance evaluation. Classical approaches include stochastic inventory models, service-level optimization, and multi-echelon inventory theory~\cite{Kok2018ATAA}. While forecasting methods such as ARIMA, exponential smoothing, and state-space models remain widely used in practice, more recent work incorporates machine learning and data-driven methods for demand forecasting, anomaly detection, and supply chain optimization~\cite{Zhu2025SustainableOIA}. Surveys on intelligent supply chain management and digital supply chains highlight the increasing integration of analytics, simulation, and AI techniques~\cite{Andaloussi2024ABLA}.

\subsection{AI and LLMs for Supply Chain Analytics}

Recent work has explored AI-based assistants and agentic systems for SCM, including conceptual analyses of AI integration and sociotechnical dynamics~\cite{ArtificialIntelligenceSCM}, renewed interest in agent-based automation~\cite{WillBotsTakeOverSC}, and domain-specific applications such as inventory replenishment~\cite{AgenticAIInventory}, sustainable automation~\cite{SustAISCM}, LLM-based multi-agent inventory control~\cite{InvAgent}, enterprise AI agents~\cite{DataBricksAgents}, and generative planning frameworks~\cite{Yin2025RethinkingSCA}. 

While these approaches demonstrate the feasibility of agent-based and LLM-driven SCM automation, they primarily focus on specific functional domains or high-level paradigms. In contrast, our work proposes a unified architecture that integrates natural-language interaction, schema-grounded data retrieval, deterministic workflows, and exploratory reasoning within a controlled multi-agent framework.

\section{Agentic System Architecture}

\begin{figure}[h]
    \centering
    \includegraphics[width=0.6\linewidth]{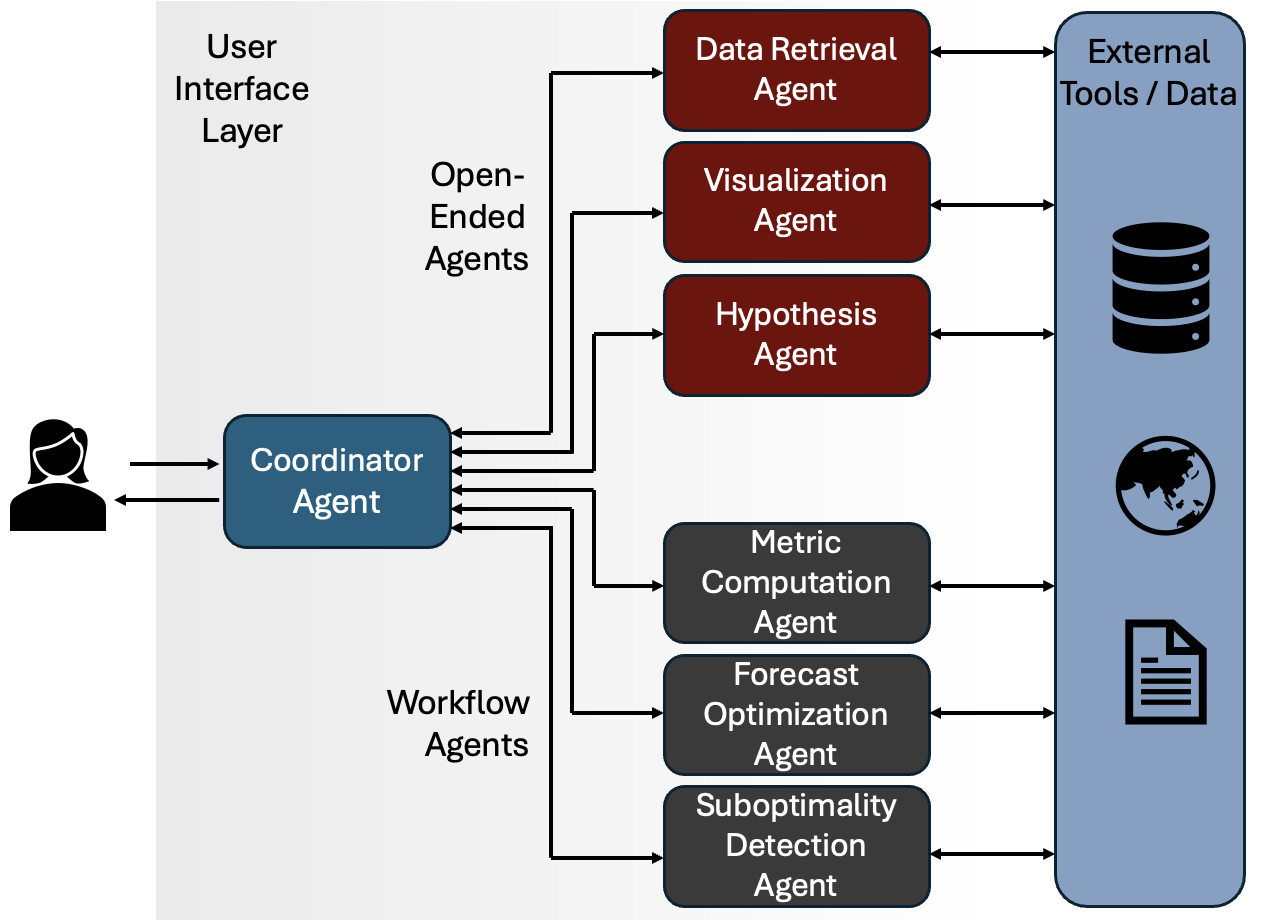}
    \caption{Overview of the agentic architecture consisting of the coordinator, open-ended and workflow agents.}
    \label{fig:architecture}
\end{figure}

The design of the proposed agentic system is guided by three primary goals driven by business necessities: usability, extensibility, and scalability.

\textbf{Usability}: Planners need intuitive access to supply chain data and analytics without requiring technical expertise in databases, query languages, or specialized tools. High usability reduces training overhead, accelerates decision-making, and ensures insights can be acted upon efficiently.

\textbf{Extensibility}: Supply chain environments evolve rapidly, with new data sources, analytical tools, functions, or agent capabilities emerging over time. An extensible system can incorporate these additions seamlessly, supporting continuous improvement and adaptation to changing business needs.

\textbf{Scalability}: As organizational usage grows, systems must handle increased query volume, larger datasets, and more complex analyses efficiently. Scalability ensures that operational costs remain manageable while maintaining performance and responsiveness under higher workloads.

We illustrate how our proposed architecture fulfill these goals in the following sections below.

% \textbf{Usability:} The system enables planners to query and analyze supply chain data through natural language, without requiring knowledge of database schema, query languages, or analytical tools. Domain knowledge, such schema mappings, KPI definitions, and standardized analytical procedures are encapsulated within specialist agents. Importantly, the architecture supports both open-ended exploratory analysis and deterministic, domain-specific workflows, allowing users to flexibly investigate issues while ensuring structured and auditable execution for critical processes.

% \textbf{Extensibility:} To lower the barrier to system evolution, routing logic, workflow defaults, parameter templates, and method-selection rules are defined primarily within agent instructions rather than hard-coded application logic. New analytical capabilities or workflow variants can therefore be introduced by modifying prompts and adding specialist agents, reducing development overhead and supporting incremental system growth.

% \textbf{Scalability:} To mitigate the context overload of monolithic agents, responsibilities are modularized across focused specialist agents with constrained tool access. By distributing reasoning across agents and interaction turns, the architecture maintains clearer role boundaries, improves interpretability, and supports larger analytical scopes without overwhelming a single model context.

\subsection{Hub-and-Spoke Multi-Agent Architecture}
Fig.~\ref{fig:architecture}~\label{subsec:multiagent_arch} illustrates the proposed system, a hub-and-spoke architecture centered on a coordinator agent. The coordinator serves as the entry point for user interaction: it interprets natural-language queries, decomposes them into sub-tasks, and delegates to specialist agents, supporting the usability of the system. In this work, we define two types of specialist agents: open-ended agents, which support flexible exploration and ad-hoc analyses, and workflow agents, which execute pre-defined, domain-specific procedures. All communication between specialists is mediated by the coordinator; specialist agents do not directly communicate with each other. This design simplifies control flow, improves auditability, and ensures that intermediate results are consistently synthesized into a coherent response. All inter-agent data exchange is mediated through explicit data artifacts rather than raw conversational context. Agents write outputs to files and return references with metadata, enabling reuse of intermediate results, improved scalability, and a clear audit trace linking outputs to specific agents.
\\
\\
\noindent\textbf{Coordinator agent}\\
The coordinator is responsible for intent recognition, task decomposition, and result synthesis. Its instructions encode patterns for common workflows, such as retrieving data and producing a visualization, or running deterministic workflows. When a request involves multiple tasks or entities (e.g., multiple metrics or SKUs), the coordinator decomposes the request into separate calls and aggregates the results before responding to the user.
\\
\\
\noindent\textbf{Open-ended agents}\\
Each specialist agent encapsulates a distinct analytical capability and is exposed to the coordinator through a predefined message schema. Rather than returning raw outputs, agents produce structured summaries that reference data artifacts and report execution status, allowing the coordinator to reason over outcomes without ingesting large contexts. We describe representative open-ended specialist agents below.
\\
\\
\noindent \textit{Data Retrieval Agent}: Generates and executes database queries over the supply chain databases. Its instructions encode relational schema knowledge and entity-mapping rules (e.g., warehouse--store and store--customer relationships). Instead of returning raw query results, it writes outputs to a data artifact and returns a reference with metadata such as row counts, column summaries, and execution status. This prevents large datasets from entering the coordinator’s context while enabling downstream reuse.
\\
\\
\noindent \textit{Visualization Agent}: Generates plots from referenced data artifacts. Given a data reference and high-level plotting intent, it loads the data, produces the visualization, and returns a plot reference with a brief description. Isolating visualization logic within a dedicated agent cleanly separates analytical computation from presentation.
\\
\\
\noindent \textit{Hypothesis Generation Agent}: Supports exploratory analysis without a predefined workflow. It can  retrieve data, perform analyses, and incorporate intermediate artifacts to produce a set of hypotheses. Each hypothesis includes a problem statement, supporting evidence, severity assessment, and confidence estimate. It complements deterministic workflows by addressing open-ended questions such as detecting data quality issues or operational anomalies.
\\
\\
\noindent\textbf{Workflow agents}\\
Workflow agents encapsulate deterministic, domain-specific procedures that follow predefined sequences of steps. Unlike open-ended agents, these agents implement structured and auditable workflows aligned with established supply chain practices. By formalizing multi-step analyses within controlled execution paths, they enable domain knowledge defined by experts to be systematically embedded and reused within the system. 

Together, these agents illustrate how deterministic, domain-safe procedures coexist with open-ended exploratory agents within the same hub-and-spoke architecture. The coordinator’s routing logic selects between flexible hypothesis-driven analysis and fixed multi-step workflows, enabling a unified yet controlled analytical framework.
\\
\\
\noindent \textit{Metrics Agent}: Computes predefined supply chain KPIs, for example: days inventory outstanding (DIO) and Fill Rate, but is generalizable to any commonly used KPIs. It accepts high-level parameters such as company, item, and date range, and returns KPI values with summary statistics. 
\\
\\
\noindent \textit{Sub-optimality Detection Agent}: Runs a workflow built based on established practices for identifying and diagnosing under-performing stores . In our implementation, the process begins with template filling to define the analysis scope (e.g., stores, SKU, date range, target service level). KPI values are computed, followed by a meet-target analysis to identify stores failing service thresholds. A skyline analysis then evaluates the DIO–fill rate trade-off to distinguish dominant and dominated stores. Additional single- and multi-store diagnostics incorporate indicators such as forecast accuracy, delivery performance, demand variability, and safety stock ratios. 
\\
\\
\noindent \textit{Forecast Optimization Agent}: Executes a deterministic re-forecasting workflow. Historical demand data are retrieved from a referenced artifact, and multiple forecasting methods are evaluated and the method with the lowest error is selected. The agent then produces new forecasts for the requested horizon and may conduct what-if simulations comparing baseline and optimized forecast scenarios. Additional details are provided in Section~\ref{subsec:case_study}
\\
\\
\noindent\textbf{Tools}\\
To successfully complete tasks, agents require a controlled degree of autonomy and access to specific execution tools. In this work, we restrict the tool-set to Python and SQL code executors, simulated enterprise databases, predefined functions for metric computation and workflow activation, a supply chain simulator, and a virtual file system.

\section{Experimental Setup}

\subsection{Experimental Testbed}

To evaluate the proposed framework, we constructed a controlled experimental environment that emulates a simplified enterprise supply chain. The modeled network consists of one supplier, one warehouse, and ten retail stores serving end customers. The system manages five distinct item types. Lead times between facilities are fixed, while customer demand for each item at each store is randomly generated to introduce variability. An enterprise-grade simulator was used to generate transactional and operational data, producing a rich dataset representative of real-world supply chain operations. The dataset is stored in a relational database to mimic real-world ERP database, including tables for store metadata, demand forecasts, item prices, inventory levels, lead times, delivery records, and related operational attributes. The database is exposed through APIs accessible to the agentic system via tool calls.

\subsection{Experiments and Tasks}\label{subsec:tasks}

We curated an initial set of 20 representative queries from a domain expert to reflect realistic planner interactions, spanning simple data retrieval, visualization requests, deterministic workflow triggers (e.g., sub-optimality detection and re-forecasting), open-ended diagnostic questions, and multi-step analytical tasks. To increase coverage, we expanded this set to 100 queries using an external LLM distinct from those used in evaluation to avoid contamination, and manually reviewed them for realism and validity. Queries are categorized into four types - \textit{analysis, data retrieval, visualization} and \textit{metric calculation} - and sampled proportionally for the reliability evaluation described in Section~\ref{subsec:eval_protocol}. Some sample queries are shown in Table~\ref{tab:sample_queries}. The agentic framework was instantiated with the specialized agents described above. We compared four configurations:
\begin{itemize}
    \item Multi-agent architecture with reasoning model
    \item Multi-agent architecture with non-reasoning model
    \item Single-agent architecture with reasoning model
    \item Single-agent architecture with non-reasoning model
\end{itemize}

\begin{table}[h]
\centering
\caption{Sample User Queries}
\begin{tabular}{p{0.95\linewidth}}
\toprule
\textbf{Sample Queries} \\
\midrule
1. What is the delivery lead time between the warehouse and store\_01 for June and July 2024? \\[0.3em]

2. Plot the actual forecast for Item2 in store\_01 from December 1, 2024 to January 31, 2025. \\[0.3em]

3. Explain the demand patterns for Item1 and Item2 at store\_01. \\[0.3em]

4. Which store had the highest demand for Item1 in April 2024? \\[0.3em]

5. What is the fill rate for Item2 in store\_01 from January 12, 2025 to January 30, 2025? \\
\bottomrule
\end{tabular}
\label{tab:sample_queries}
\end{table}

For the single-agent baseline, we consolidated the prompts of all specialist agents into a unified prompt with minimal semantic modification, ensuring a fair comparison. The reasoning configuration uses \texttt{o4-mini}, while the non-reasoning configuration uses \texttt{GPT-4.1}. Each configuration was evaluated on the full set of 100 queries across five independent runs. All experiments were implemented using OpenAI’s Agent SDK.

\subsection{Evaluation Protocol}\label{subsec:eval_protocol}

To enable scalable and consistent evaluation, we adopt an LLM-as-judge framework~\cite{llm-as-judge}, which scores each query--response pair along three dimensions: (i) whether the response addresses the user’s intent, (ii) completeness, and (iii) relevance. Scores are assigned on a discrete scale of $0$ (fail), $0.5$ (partial), and $1$ (success), accompanied by a brief justification. Evaluation was performed using GPT-5.2 to ensure independence from the models used for query expansion and system execution.

To validate the reliability of automated scoring, we sampled 20 expert-generated queries, covering the four task categories and had a domain expert manually assess correctness. Out of 20 queries, the expert fully agreed with the LLM-as-a-judge in 17 cases, partially agreed in 2, and disagreed in 1, demonstrating strong alignment. While responses are often data-dependent and the LLM may misjudge plausible but incorrect answers, this high agreement suggests that the automated judge provides a reasonable proxy for factual correctness. This validation supports the interpretation of subsequent results in the following section.

\section{Results and Discussion}

\subsection{Experimental Results}

To compare architectural and model effects, Table~\ref{tab:combined_results} reports mean scores averaged across five runs. We observed that the choice of underlying model significantly influences performance: reasoning models achieve scores above 90\%, while non-reasoning models achieve  $\approx65\%$.

Interestingly, architectural effects differ by model type. For reasoning models, the multi-agent configuration shows a slight decrease in performance compared to the single-agent configuration, with similar variance across runs. Conversely, for non-reasoning models, the multi-agent architecture performs slightly better with a significantly lower overall variance than the single-agent baseline. In general, reasoning models exhibit more stable performance across experiments.

We hypothesize that the single-agent configuration performs better when paired with a reasoning model because all prompts are concatenated, giving the LLM full visibility of every agent role and capability. When combined with stronger reasoning ability, this global context may allow the model to internally coordinate tasks more effectively. In contrast, within the multi-agent system, each agent operates with a limited and role-specific context. While this structured separation can improve clarity, it may restrict the benefits of stronger reasoning models that could otherwise leverage broader system-level information. However, when paired with non-reasoning models, the decentralization and explicit division of responsibilities appear to compensate for weaker internal reasoning, leading to modest performance gains.

Table~\ref{tab:combined_results} reports input token usage. As expected, reasoning models consume approximately at least 1.5-2.5X as many tokens as non-reasoning models due to the generation of intermediate reasoning traces. More importantly, the single-agent architecture consumes roughly four times more tokens than the multi-agent architecture. This occurs because the single-agent prompt contains the full specification of all system roles and workflows in every interaction, whereas the multi-agent design activates only the relevant prompts required for a given task.

From a scalability perspective, this difference reflects a fundamental trade-off between global context aggregation and modular decomposition. Since LLMs operate within fixed context windows, continuously expanding a single prompt with all functional instructions increases context load and may introduce interference among unrelated tasks. The multi-agent architecture distributes responsibilities across specialized agents, ensuring that only task-relevant information is exposed at each step. This provides a systematic mechanism for controlling context growth and improving token efficiency. While this may result in a slight drop in peak accuracy under strong reasoning models, it represents a more scalable and principled system design. Nonetheless, the multi-agent architecture has limitations. With current LLMs, a single unified agent that has full observability of all tasks may achieve slightly higher performance when paired with strong reasoning capability. Additionally, multi-agent systems introduce communication overhead among agents, which can increase response latency.

\begin{table}[h]
\centering
\caption{Performance and Relative Input Token Usage (Lowest = 1$\times$). NR denotes non-reasoning models.}
\begin{tabular}{lccc}
\toprule
\textbf{Model} & \textbf{Setup} & \textbf{Mean Score} & \textbf{Tokens (Rel.)} \\
\midrule
GPT-4.1 & Single-Agent (NR) & $0.655 \pm 0.027$ & $5.5\times$ \\
GPT-4.1 & Multi-Agent (NR)  & $0.661 \pm 0.018$ & $1.0\times$ \\
o4-mini & Single-Agent & $0.932 \pm 0.012$ & $8.4\times$ \\
o4-mini & Multi-Agent  & $0.904 \pm 0.011$ & $2.5\times$ \\
\bottomrule
\end{tabular}
\label{tab:combined_results}
\end{table}

\subsection{Analysis of failures}

Table~\ref{tab:distribution_results} shows the distribution of scores across configurations. The reasoning models exhibit a strong concentration of successful responses, while the non-reasoning models display a more even distribution between successful and partially correct responses. This suggests that reasoning capability primarily affects the system’s ability to consistently complete tasks rather than partially address them.

To analyze performance across task types, we use the query categorization introduced in Section~\ref{subsec:tasks}. The results indicate that the largest performance gap between reasoning and non-reasoning models occurs in data retrieval and visualization tasks, whereas structured metric calculation tasks exhibit a smaller gap. Interestingly, the multi-agent configuration can improve performance for non-reasoning models in some categories, highlighting the benefits of modular delegation even when reasoning capability is limited.

Closer inspection of individual failures (see Table~\ref{tab:sample_failures}) reveals that non-reasoning models frequently fail during data retrieval due to technical execution errors rather than high-level reasoning mistakes. Many failures stem from issues such as incorrect date serialization, improper handling of date formatting, missing tables, or misconfigured data source connections. These errors often occur when constructing or executing SQL queries. Since visualization and downstream analytical tasks depend on successful data retrieval, such low-level execution issues propagate, leading to compounded failures across the workflow.

Overall, these results suggest that reasoning capability and architectural design jointly shape system robustness in schema-sensitive and multi-step tasks. Stronger reasoning models improve reliability in query construction and execution, particularly for data retrieval and visualization. At the same time, the multi-agent architecture contributes modular task decomposition, controlled workflow execution, and improved token efficiency by activating only the necessary components for each query. Together, these factors enhance not only performance stability but also scalability and cost-effectiveness.

\begin{table}[h]
\centering
\caption{Outcome Distribution Across Configurations}
\begin{tabular}{llccc}
\toprule
\textbf{Model} & \textbf{Architecture} & \textbf{Fail (\%)} & \textbf{Partial (\%)} & \textbf{Success (\%)} \\
\midrule
GPT-4.1 & Single-Agent  & 8.4  & 52.2 & 39.4 \\
GPT-4.1 & Multi-Agent   & 10.6 & 46.6 & 42.8 \\
o4-mini & Single-Agent  & 1.0  & 11.6 & 87.4 \\
o4-mini & Multi-Agent   & 1.6  & 16.0 & 82.4 \\
\bottomrule
\end{tabular}
\label{tab:distribution_results}
\end{table}

\begin{table}[h]
\centering
\caption{Performance by Task Category (Mean Score, Single-/Multi-Agent)}
\begin{tabular}{lccc}
\toprule
\textbf{Task Category} & \textbf{o4-mini} & \textbf{GPT-4.1} \\
 & (Single / Multi) & (Single / Multi) \\
\midrule
Analyze or Summarize Results & 0.96 / 0.92 & 0.72 / 0.73 \\
Retrieve Data                & 0.90 / 0.88 & 0.55 / 0.54 \\
Visualize Results            & 0.99 / 0.94 & 0.50 / 0.47 \\
Calculate Metrics            & 0.85 / 0.90 & 0.77 / 0.86 \\
\midrule
\textbf{Average}         & 0.94 / 0.91 & 0.66 / 0.66 \\
\bottomrule
\end{tabular}
\label{tab:category_results_singlecol}
\end{table}

% \begin{table*}[h]
% \centering
% \caption{Performance by Task Category (Mean Score)}
% \begin{tabular}{lccccc}
% \toprule
% \textbf{Task Category} & \textbf{Count} & \textbf{o4-mini} & \textbf{o4-mini} & \textbf{GPT-4.1} & \textbf{GPT-4.1} \\
%  &  & \textbf{Single-Agent} & \textbf{Multi-Agent} & \textbf{Single-Agent} & \textbf{Multi-Agent} \\
% \midrule
% Analyze or Summarize Results & 53 & 0.959 & 0.923 & 0.717 & 0.725 \\
% Retrieve Data                & 24 & 0.904 & 0.875 & 0.554 & 0.542 \\
% Visualize Results            & 13 & 0.992 & 0.938 & 0.500 & 0.469 \\
% Calculate Metrics            & 10 & 0.850 & 0.900 & 0.770 & 0.860 \\
% \midrule
% \textbf{Grand Total}         & 100 & 0.939 & 0.911 & 0.655 & 0.661 \\
% \bottomrule
% \end{tabular}
% \label{tab:category_results}
% \end{table*}

\begin{table}[h]
\centering
\caption{Sample Failure Responses}
\begin{tabular}{p{0.95\linewidth}}
\toprule
\textbf{Representative Sample of Failure Responses} \\
\midrule
1. It appears there was an error retrieving the delivery lead time data due to a technical issue with date formatting in the system. The delivery and shipment dates for warehouse to store\_01 deliveries for June and July 2024 could not be fetched. \\[0.6em]

2. There was an error retrieving the data due to a technical issue with date serialization in the query process. This issue relates to how the database handles date fields for output formatting rather than the underlying database structure. \\[0.6em]

3. The system was unable to retrieve the demand data for Item1 and Item2 at store\_01 due to a technical issue with the data source. Specifically, date fields could not be processed and the required file for visualization could not be found. \\[0.6em]

4. An error occurred while retrieving delivery data for store\_01 and store\_02, likely because the deliveries table does not exist in the current database or the data source connection is misconfigured. \\[0.6em]

5. There was an error retrieving inbound shipment data due to a date serialization issue during SQL execution, indicating a technical problem within the tool. \\
\bottomrule
\end{tabular}
\label{tab:sample_failures}
\end{table}

\subsection{Case study: Hybrid Workflows}\label{subsec:case_study}

Next, we present a case study illustrating how domain knowledge is integrated into the agentic system, and how the system seamlessly combine open-ended data-retrieval agents with domain-informed workflow agents to retrieve relevant data and autonomously trigger appropriate analyses.

\begin{figure}[h]
    \centering
    \includegraphics[width=0.6\linewidth]{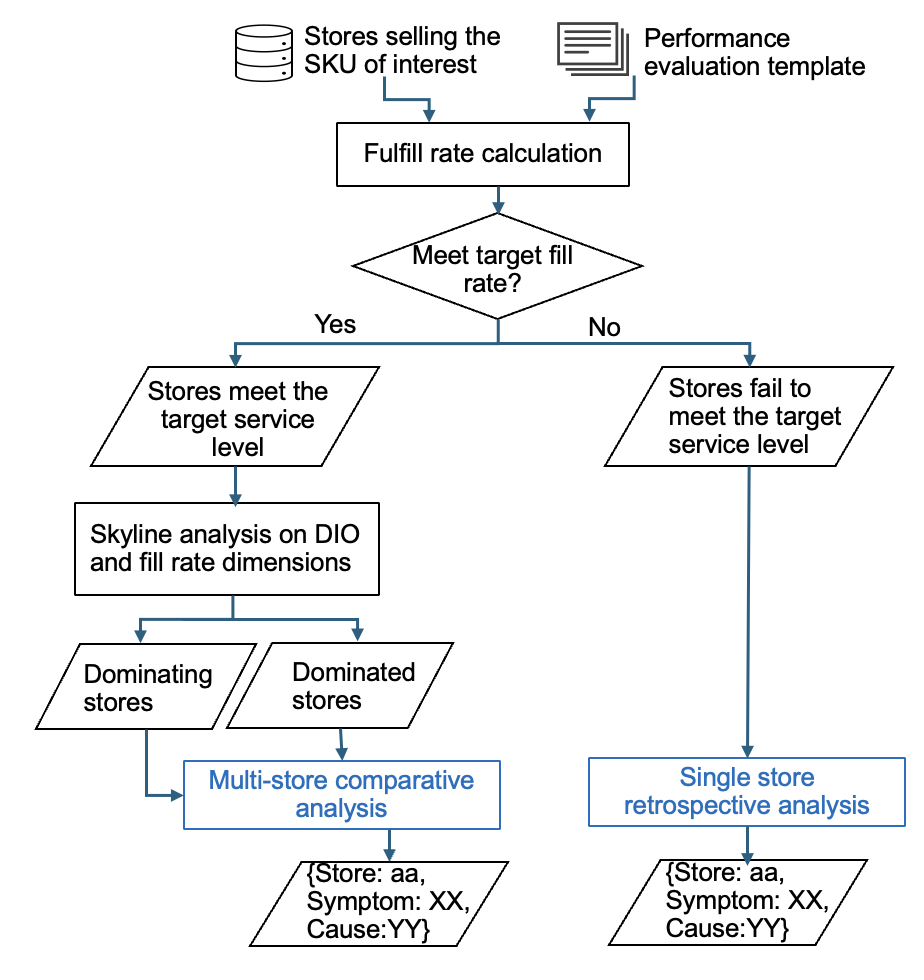}
    \caption{Workflow for store sub-optimality detection.}
    \label{fig:suboptimality_detection_workflow}
\end{figure}

Using the same experimental testbed, we consider a sub-optimality detection workflow (Figure~\ref{fig:suboptimality_detection_workflow}). It begins with the planner identifying a set of candidate stores and SKUs suspected to be suboptimal. The planner specifies a performance evaluation template, including the time horizon, SKUs of interest, KPIs used for sub-optimality assessment, and KPI computation cycles. We assume that achieving a specified customer service level is the primary objective. Stores are first classified based on whether they meet a target fulfillment rate. Those meeting the targets then undergo skyline (Pareto frontier) analysis along selected KPI dimensions (e.g., DIO and fill rate) to identify dominant and dominated stores.
Subsequently, a comparative analysis is conducted in which the operating parameters of dominated stores are assessed against those of dominant stores to identify potential root causes, including delivery lead time, demand forecast accuracy, and safety stock levels. Stores that fail to meet the fulfillment threshold undergo a single-store retrospective analysis to detect potential distribution shifts in demand, delivery time anomalies, or forecast errors. The final output is a set of suboptimal stores together with their associated symptoms and likely root causes. This workflow is inherently complex, comprising of nested and conditionally triggered sub-workflows. Traditionally, executing such a process would require coordination between planners and data analysts, with each step involving separate data retrieval operations and specialized analyses.

We demonstrate that this entire process can be embedded within the agentic architecture, thereby reducing reliance on data analysts and accelerating data driven decision-making. A natural-language interface first parses the planner’s request to identify relevant stores and SKUs. The coordinator agent then invokes the data-retrieval agent to dynamically populate the performance template, followed by deterministic execution of the appropriate analytical workflows. While the workflow logic is structured and domain-defined, the data retrieval and instantiation are dynamically driven by the planner’s query. The final output provides not only the identified suboptimal stores but also interpretable diagnostic evidence, thereby supporting traceability and auditability. Text box~\ref{fig:suboptimality_output} illustrates a sample query and the corresponding output of the agentic system.

\begin{tcolorbox}[title=Query: Perform suboptimal analysis of Item1 from 2025-02-15 to 2025-03-15 using 90\% target fill rate),  breakable]
\small
\label{fig:suboptimality_output}

\textbf{1. Stores Meeting Target Fill Rate} \\
store\_01, store\_02, store\_03, store\_04, store\_05, store\_06, store\_07, store\_08, store\_09, store\_10

\vspace{0.5em}
\textbf{2. Stores Not Meeting Target} \\
None

\vspace{0.5em}
\textbf{3. Dominating Stores} \\
store\_01 (fill rate: 1.00, DIO: 0) \\
store\_08 (fill rate: 1.00, DIO: 0)

\vspace{0.5em}
\textbf{4. Dominated Stores} \\
store\_03, store\_05, store\_07, store\_09, store\_10 (fill rate: 0.99, DIO: 0) \\
store\_02 (fill rate: 0.98, DIO: 0) \\
store\_04, store\_06 (fill rate: 0.99, DIO: 3)

\vspace{0.5em}
\textbf{5. Skyline Plot} \\
See saved figure at skyline.png

\vspace{0.5em}
\textbf{6. Single-Store Diagnostics (Dominated Stores)} \\
MAPE range: 0.05 -- 0.07 \\
Primary symptom: lower Fill Rate or higher DIO \\
Cause: No definitive root cause identified within predefined indicators

\vspace{0.5em}
\textbf{7. Multi-Store Comparative Analysis} \\
Contributing factors are:
\begin{itemize}
    \item Poor forecast accuracy (forecast accuracy ratio $>$ 1.1)
    \item Excessive safety stock (ratio up to 6.0)
    \item Demand variability differences
\end{itemize}

\textbf{Recommendation:} Consider forecast re-optimization for stores exhibiting poor forecast accuracy or excessive safety stock (e.g., store\_02, store\_05, store\_06, store\_09, store\_10).
\end{tcolorbox}

Building on the sub-optimality results, suppose the planner seeks to improve the performance of a specific store, e.g., $store\_06$. This triggers a second workflow agent: the forecast optimization agent, which has access to multiple forecasting algorithms. The agent autonomously retrieves the relevant historical data for the selected store and SKU, evaluates several candidate models, and selects the best-performing method. In our implementation, the available models include Gaussian Process regression, Exponential Smoothing, and ARIMA. The selected model is then fitted and deployed to generate an improved forecast.

\begin{figure}
    \centering
    \includegraphics[width=1\linewidth]{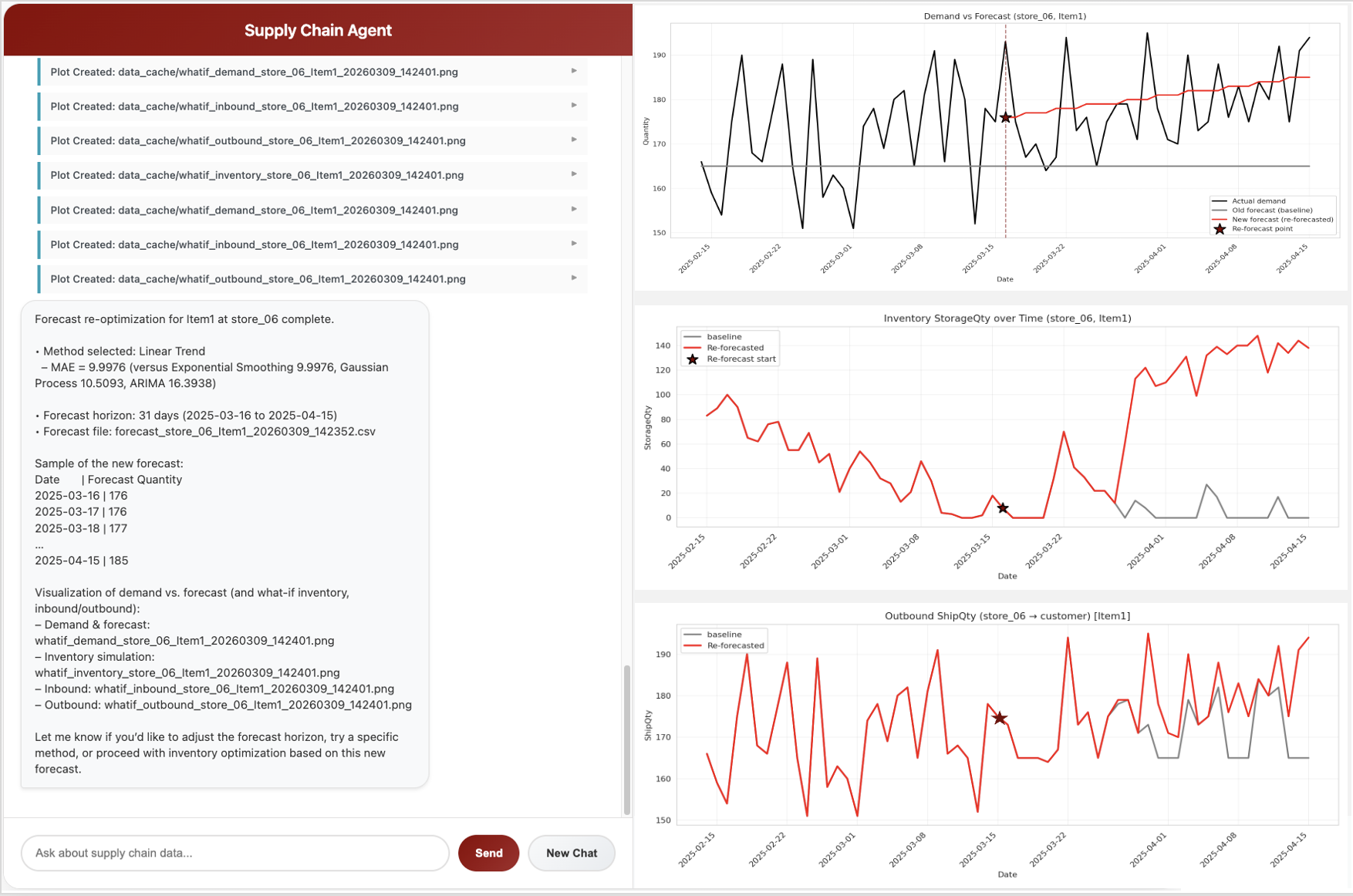}
    \caption{The agentic system's user interface with what-if simulation results comparing baseline and optimized forecast for the sub-optimality detection and demand re-forecasting case study. }
    \label{fig:forecast_optimization_output}
\end{figure}

To assess downstream impact, we integrate our agentic system with a supply chain digital twin and compare the supply chain performance under the original and optimized forecasts. Fig.~\ref{fig:forecast_optimization_output} presents the improved forecast (top plot) alongside the resulting inventory (middle plot) and the outbound shipment trajectories for $store\_06$ (bottom plot). The optimized forecast (shown in red) more closely tracks demand trends than the original forecast. Correspondingly, simulation results indicate sustained positive inventory levels and increased outbound shipment quantities, mitigating the stock-out scenario observed under the baseline forecast. Together, these examples demonstrate how deterministic workflows can be integrated within an agentic architecture. By combining pre-defined procedures with dynamic data retrieval and model selection, the system enables an efficient and auditable supply chain management interface.

% \begin{figure}[h]
% \centering

% \begin{subfigure}{0.9\linewidth}
%     \centering
%     \includegraphics[width=\linewidth]{whatif_demand.png}
%     \caption{Demand Forecast Comparison}
% \end{subfigure}

% \vspace{0.6cm}

% \begin{subfigure}{0.9\linewidth}
%     \centering
%     \includegraphics[width=\linewidth]{whatif_inventory.png}
%     \caption{Inventory Trajectory}
% \end{subfigure}

% \vspace{0.6cm}

% \begin{subfigure}{0.9\linewidth}
%     \centering
%     \includegraphics[width=\linewidth]{whatif_outbound.png}
%     \caption{Outbound Shipment Quantity}
% \end{subfigure}

% \caption{What-if simulation results comparing baseline and optimized forecast scenarios.}
% \label{fig:forecast_optimization_output}
% \end{figure}

% \subsection{Hypothesis Generation}

\section{Conclusion}

This paper presented a multi-agent architecture for natural-language supply chain analytics that balances flexibility with deterministic control. The system enables planners to query and analyze supply chain data via natural language without schema or tool expertise and evaluation showed the propose architecture's performance is competitive with a single-agent baseline with roughly 4X lower token usage, improving scalability and cost. A key contribution is hybrid support for open-ended exploration and domain-specific workflows, in addition to insights from the failure analysis. We also provided a case study that illustrated tight integration of deterministic workflows with dynamic data retrieval and model selection. Future work will integrate additional workflows and tools to cover more use cases (e.g., inventory transhipment), improve multi-agent performance through advanced prompt strategies, and incorporate long-context management. This work illustrates how agentic architectures can serve as a practical foundation for scalable, extensible decision-support systems in complex enterprise domains.

\bibliographystyle{unsrtnat}
\bibliography{references}  %%% Uncomment this line and comment out the ``thebibliography'' section below to use the external .bib file (using bibtex) .

\end{document}